\documentclass{article}
\usepackage{geometry}
\usepackage{graphicx}
\usepackage{authblk}
\usepackage[square,numbers]{natbib}
\usepackage{hyperref} 
\hypersetup{
    colorlinks=true,
    linkcolor=blue,
    filecolor=magenta,      
    urlcolor=cyan,
    pdftitle={Overleaf Example},
    pdfpagemode=FullScreen,
}

\title{Evolutionary Algorithms Simulating Molecular Evolution:\\ A New Field Proposal}
\author[1]{James S. L. Browning, Jr.\thanks{jlb0181@auburn.edu}}
\author[1]{Daniel R. Tauritz\thanks{drt0015@auburn.edu}}
\author[2]{John Beckmann\thanks{beckmann@auburn.edu}}
\affil[1]{Samuel Ginn College of Engineering, Auburn University, 3101 Shelby Center, Auburn, AL 36849-5347}
\affil[2]{College of Agriculture, Auburn University, 301 Funchess Hall, Auburn, AL 36845}
\date{\today}

\begin{document}

\maketitle

\begin{center}
\includegraphics[width=.25\textwidth]{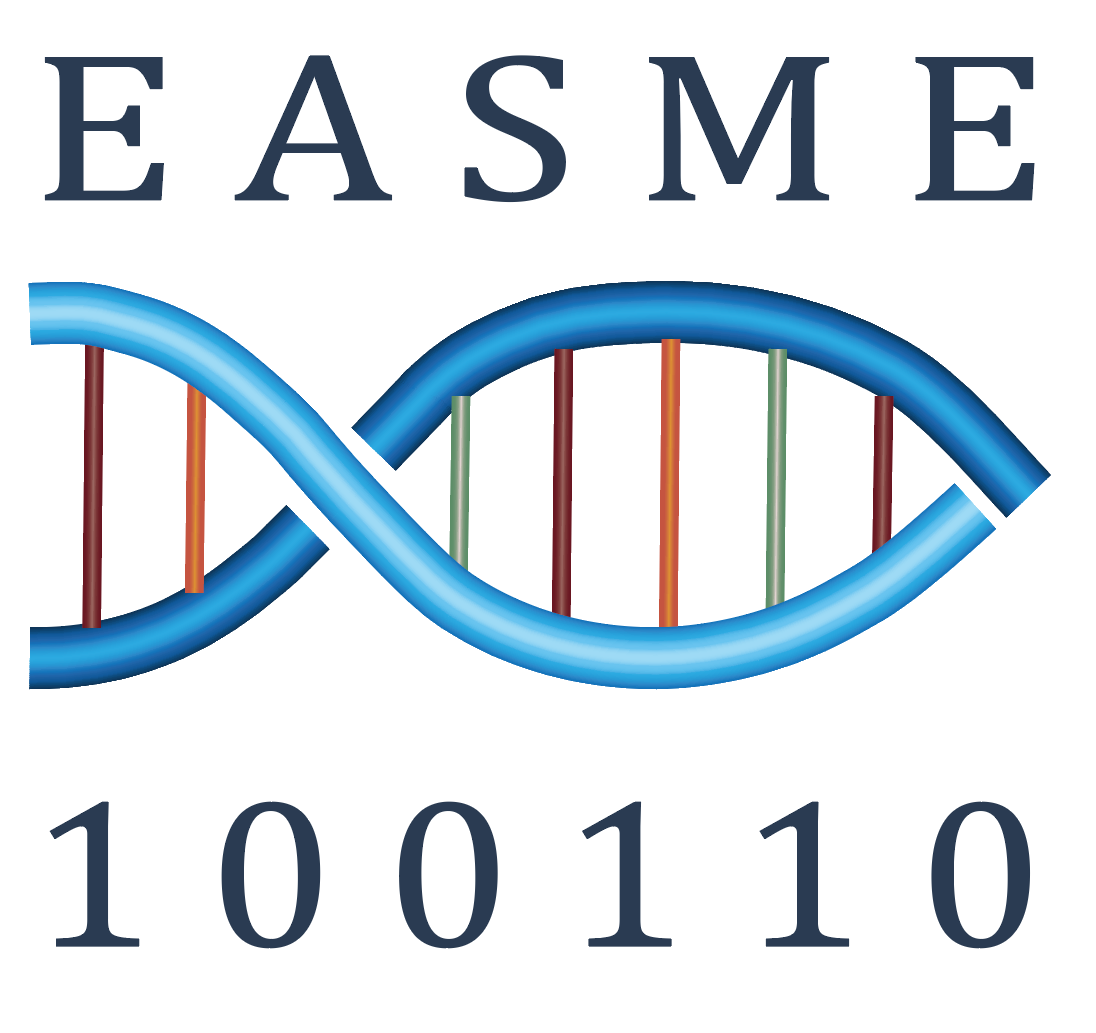}
\end{center}

\section{Key Points}

\begin{enumerate}
    \item The set of proteins produced by nature is minuscule compared to the search space of all possible proteins.
    \item AI algorithms capable of efficiently exploring this search space are now emerging.
    \item We propose a new AI framework, driven by evolutionary algorithms, capable of searching this space.
    \item Biotechnological synthesis and screening methodologies are proposed to validate the searches.
    \item The biotechnological applications of this project, were it successful, would be almost limitless.
\end{enumerate}

\section{Abstract}

The genetic blueprint for the essential functions of life is encoded in DNA, which is translated into proteins — the engines driving most of our metabolic processes. Recent advancements in genome sequencing have unveiled a vast diversity of protein families, but compared to the massive search space of all possible amino acid sequences, the set of known functional families is minimal. One could say nature has a limited protein “vocabulary.” A major question for computational biologists, therefore, is whether this vocabulary can be expanded to include useful proteins that went extinct long ago, or have never evolved (yet). By merging evolutionary algorithms, machine learning, and bioinformatics, we can develop highly-customized “designer proteins.” We dub the new sub-field of computational evolution which employs evolutionary algorithms with DNA string representations, biologically-accurate molecular evolution, and bioinformatics-informed fitness functions, Evolutionary Algorithms Simulating Molecular Evolution (EASME).

\subsection{Keywords}

Artificial Intelligence, Machine Learning, Evolutionary Algorithms, Computational Evolution, Computational Biology, Bioinformatics, Genetic Programming, Molecular Evolution, Protein Folding, Biotechnology, Synthetic Biology, Agriculture

\section{Antecedents of EASME}
\label{sec:EASME-antecedents}

Evolutionary algorithms (EAs), as the name suggests, model the natural process of evolution: with selection acting on variation where fitter individuals, i.e., individuals better adapted to their environment, have a higher chance of reproducing and surviving, thus increasing the odds of their genes proliferating within the population's gene pool. Natural sexual reproduction capitalizes on recombination between individuals and stochastic mutation to generate new alleles, thus diversifying the gene pool over many generations. By mimicking this process \textit{in silico}, EAs have proven to be powerful engineering problem solvers, outperforming human experts on many optimization and design problems, frequently through unintuitive -- or even counterintutitive -- solution approaches. In 2006, Wolfgang Banzhaf and a team of eight other experts in the fields of artificial evolution and evolutionary biology, proposed a bold new idea -- that progress in the field had slowed, but that it could continue to improve by incorporating more of the nuances of the natural world:

\begin{quote}
    We propose a richer paradigm for algorithms inspired by evolution, which we call computational evolution (CE)… Our hope is that CE will enable computational scientists to address new, difficult problems, including those of interest to natural scientists, and that the more sophisticated evolutionary modelling in CE will be directly useful to evolutionary biologists and ecologists as a basis for simulations. We also hope to promote a more sophisticated dialogue between computational and natural scientists about evolution~\cite{Banzhaf2006}.
\end{quote}

Banzhaf’s central thesis in his proposal of CE was that artificial evolution has been limited by a disconnect from the real-world mechanisms that inspired it. EASME, at its core, is an attempt to bridge that gap at the level of molecular evolution, exploring biology (particularly proteomics) with as much granularity and precision as is possible. While artificial evolution purposely abstracts away from nature, drawing only on what appears to be useful to computational problem solving, EASME embraces and attempts to model the full complexity of molecular evolution, in an attempt to break it down to its primitives.

This intersection of fields remains under-explored, and we are confident there are meaningful discoveries to be made here. For example, imagine what could be accomplished if the best minds in computational biology focused their attention on artificially optimizing just one key enzyme in photosynthesis, then translated their results into \textit{in vivo} tests on transgenic plants. Doing so might enhance agricultural production worldwide. Thus, pairing bioinformatics with EAs is an exciting possibility.

With its focus on solving biological problems, it may be that EASME is not exclusively a subset of computational evolution — Banzhaf’s proposal seemed more focused on the idea that incorporating more elements of biology would help artificial evolution solve \textit{engineering} problems:

\begin{quote}
    One example of a question that CE could address in future is whether it is possible to construct a program that functions like an organism, with interacting software objects (analogues of cells) that collectively perform a global function such as providing operating system services (such as file management), with the ability to respond gracefully to demand and damage (analogously to homeostasis).
\end{quote}

However, the potential of an even closer crossover with biology was mentioned:

\begin{quote}
    Pursuing our proposal, even imperfectly, will help to close the loop between biology and computing, for the benefit of both: CE could point to a new way of computing and a new way of understanding the living world, including how evolution works.
\end{quote}

Closing this loop at the molecular level is our goal by simulating DNA evolution in EASME projects, with fitness evaluations analyzing functional capabilities of those encoded proteins through de-novo folding and bioinformatic algorithms.

\section{Defining the Problem}

\begin{figure}[ht]
    \centering
    \includegraphics[width=.625\textwidth]{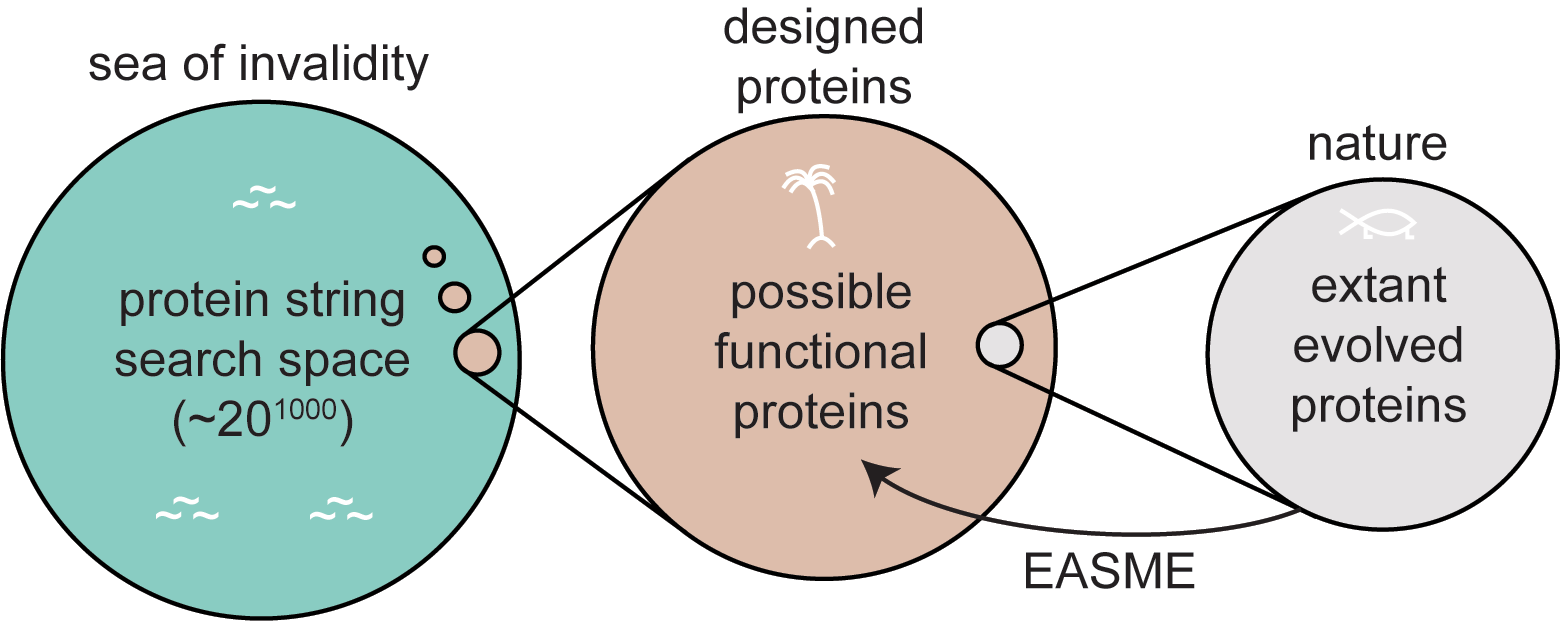}
    \caption{The search space of proteins — a vast sea of invalidity contains a handful of islands containing useful, functional proteins, only a small subset of which have likely been evolved by nature.}
    \label{fig:islands}
\end{figure}

Proteomics --- the study of proteins --- is a primary focus point for the EASME field. Protein strings are sentences written with an alphabet of 20 amino acids. Many of these amino acid strings exceed 1,000 characters in length, thus the search space of possible protein strings is unfathomably vast. Most combinations of amino acids would do absolutely nothing and would be unstable. Thus the search space of possible protein configurations is a vast “sea of invalidity.” Within that sea exists a tiny archipelago of functional proteins, and only a small region of one of those islands is occupied by the proteins that actually evolved and remain extant today (see Figure~\ref{fig:islands}). EASME aims to expand the set of extant proteins by colonizing new islands in the sea of invalidity, yielding functional protein strings that can later be produced and analyzed in a lab.

Natural evolution is driven by trial and error -- it wanders the search space and stumbles upon useful changes by chance, then refines them over generations. Importantly, the path that natural evolution traversed to arrive at any given local optima is not the only possible path, and there is no reason to conclude that what happened to evolve by nature is \textit{the} global optima for a desired function. Perhaps there are hundreds, or even thousands of Pareto optimal protein solutions for any given protein that have never been discovered, or were sporadically lost in evolutionary history. Furthermore, evolution is still ongoing, and there remain many optima yet to be found; with all the computing power now at our disposal, we believe we can speed up novel protein generation. We envision a future where whole databases of proteins could be created and mined for valuable functions through bioassay. We argue that all the tools required for this endeavour are currently available.

The EASME model employs CE to achieve this end. Recent advancements in ML have led many to claim that complex biological problems, most notably protein folding, are finally “solved.” With this being the case, one may wonder why our focus remains on CE. Ultimately, ML does have a place in the pursuit of EASME, but our reasoning for focusing on CE is twofold — one, we agree with Banzhaf that models \textit{drawing on} evolution would be the best way to study and harness evolution; and two, machine learning falls short in a key area.

\section{Where Machine Learning Falls Short}
\label{sec:ML-shortcomings}
In truth, the advances of ML have not yielded a fundamental understanding of \textit{de novo} protein folding — of the \textit{language} of proteins — as was discussed at length in “De novo protein folding on computers. Benefits and challenges” by Barry Robson~\cite{ROBSON2022105292}:

\begin{quote}
    …AlphaFold does not solve, or seek to solve, the folding problem. It “reasons” from what is ultimately biological data, not from fundamental laws of chemical physics.
\end{quote}

Importantly, ML models will always be limited by their training sets, which are restricted to the archipelago of extant functional proteins (and usually represent some smaller subset of those that have been structurally defined by X-ray crystallography or cryogenic electron microscopy). These minimal data sets are also biased in that they favor proteins from eukaryotes that are involved in health issues. Importantly, the training data for ML models can only be as large as the minimal set of proteins bestowed upon us by nature. Unless ML processes learn intrinsic grammar rules and folding patterns dictated by biophysics, it will remain difficult to generate true novelty in structure and function. The current state of many areas of computer science could be defined similarly — while astounding advancements have surely been made in the field of deep learning, and this has yielded very impressive results in some cases, these results are ultimately facsimiles of \textit{what} is, not a true understanding of \textit{why} it is (or what it \textit{could} be).

\section{Where Evolutionary Algorithms Can Help}

When it comes to uncovering the \textit{why} of something, evolutionary computation holds a unique advantage. This was most succinctly observed in D’Angelo \textit{et al.}’s 2023 paper “Identifying patterns in multiple biomarkers to diagnose diabetic foot using an explainable genetic programming-based approach,” in which (as the title suggests) genetic programming — a type of EA employing complex representations — was used to diagnose diabetic foot~\cite{DANGELO2023138}. Not only did the team’s genetic programming approach \textit{outperform} ML, but the decisions the program produced were easily comprehensible by its human operators. Whereas ML often produces “black boxes,” EAs (and their various bio-mimetic kin) build and evolve solutions using simple, predefined primitives. All AI algorithms have their own pros and cons, of course, and can approach unique problems in different ways. We merely argue that the best path to discovering new proteins is likely a hybrid AI approach, where an EA sits at base of the program, acting as the engine that drives and selects for novelty in the context of protein grammar rules and bioinformatic insights.

\section{Evolutionary Algorithms Simulating Molecular Evolution}

EASME is short for \textbf{evolutionary algorithms simulating molecular evolution}. To summarize what has been discussed so far, Banzhaf \textit{et al.} proposed a research agenda for the new field of CE in 2004, suggesting that our modern understanding of evolution be applied to creating more performant EAs. We have identified a very valuable use case for this line of thinking, that being the development of truly novel functional proteins, which ML models would struggle to produce on their own.

As computing power continues to increase, complex simulations of evolving biochemical systems are approaching feasibility, even on desktop computers. This has been demonstrated most recently in ``Modeling the emergence of \textit{Wolbachia} toxin antidote protein functions with an evolutionary algorithm''~\cite{10.3389/fmicb.2023.1116766}, a project in which researchers drove forward evolution of two specific, interacting proteins involved in cytoplasmic incompatibility, and observed the speed, location, and addition of novel functions that can be mapped to phenotypes. In another article, “Molecular dynamics simulation of an entire cell” by Stevens \textit{et al.}, an entire minimal cell was simulated \textit{in silico} (stopping just short of full simulation of metabolic activity, which still proved difficult)~\cite{10.3389/fchem.2023.1106495}. With the computing power to simulate complex biochemical systems now at our disposal, and EAs that model the same process that \textit{produces} those biomolecules, the stage is now set for the EASME field to bloom.

\begin{figure}[ht]
    \centering
    \includegraphics[width=.625\textwidth]{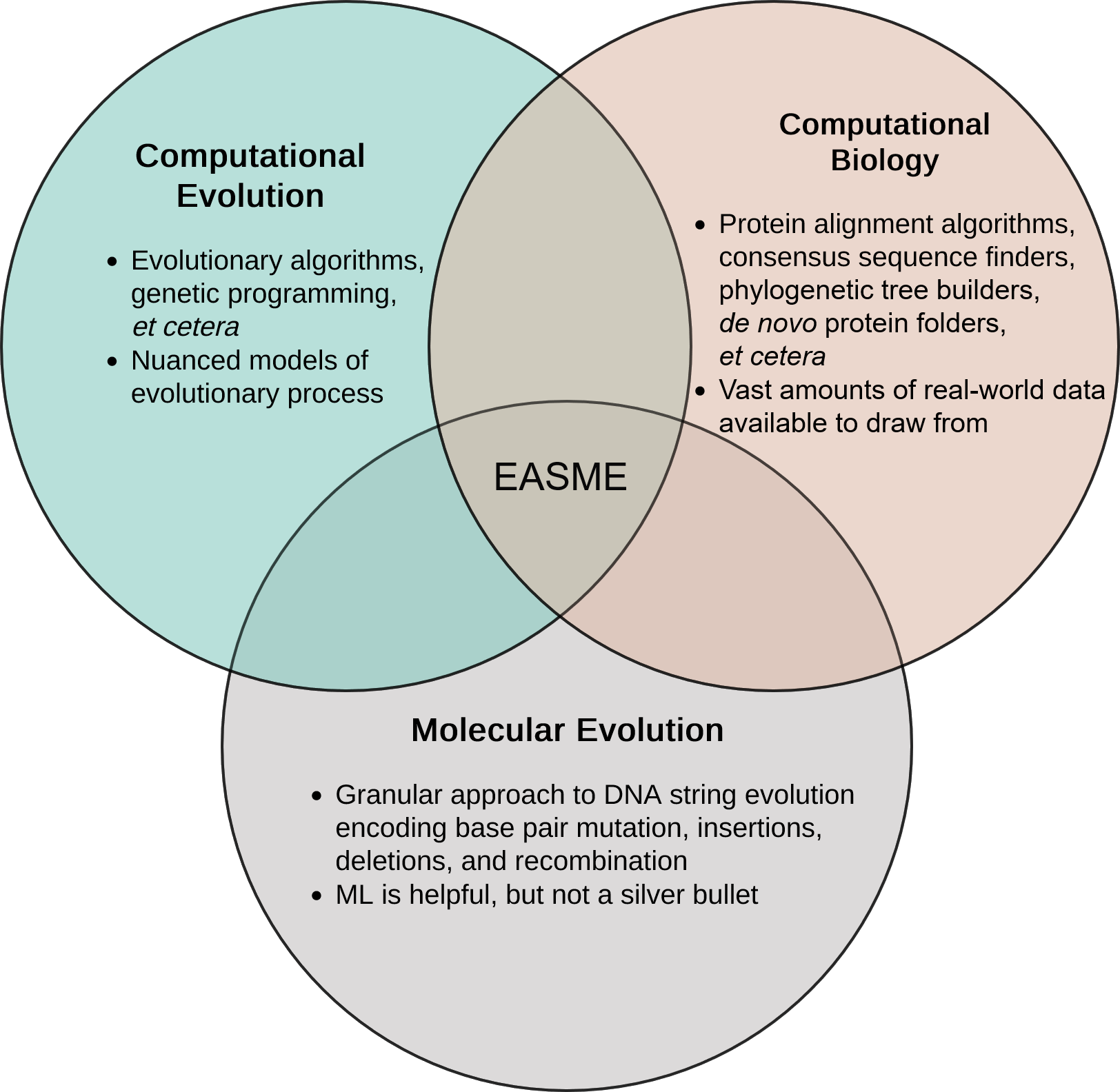}
    \caption{EASME exists at the intersection of computational evolution, computational biology, and molecular evolution, drawing on the strengths of all three to solve problems.}
    \label{fig:venn}
\end{figure}

Through the continued development of this burgeoning field — leveraging the biomimicry of EAs (in particular genetic programming), ML, and state-of-the-art hardware — we feel uncovering grammar structures that define the fundamental languages of our biology is now possible (see Figure~\ref{fig:venn}). While this work will undoubtedly be complex, we are confident that utilizing all the AI tools now at our disposal will nonetheless make it feasible.

One promise of the EASME field, a possibility our team is currently exploring, is a deeper look into the origins of life, the formation of the first self-replicating biomolecules, and the genesis of the first self-replicating cells. With approximations of protein stability and \textit{de novo} folding structure now feasible on common hardware, in combination with precisely defined fitness functions and protein grammar rules, these features could be used as the objective functions for an EA, and a novel \textit{in silico} “tree of life” could be  grown and regrown for the first proteins on Earth.

\section{Building and Testing the EASME Algorithm}

\begin{figure}[ht]
    \centering
    \includegraphics[width=.625\textwidth]{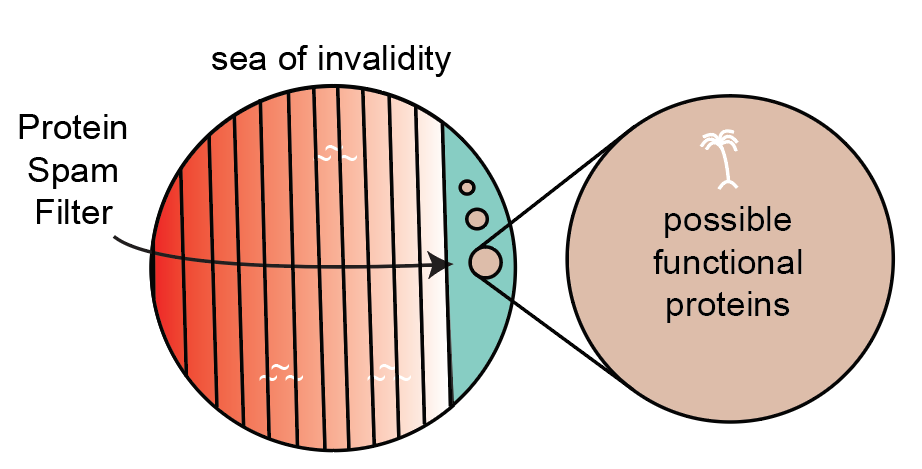}
    \caption{Intelligent application of ML can help constrain the search space of proteins.}
    \label{fig:filter}
\end{figure}

Our hypothesis is that an EASME model would be composed of several algorithmic components. At the base of EASME would be an EA that would instantiate diverse populations of DNA sequences. The base EA would evolve and mutate genes the same way they naturally evolve — by point mutations, deletions, insertions, and recombinations. Fitness of encoded proteins would be determined by a multifaceted fitness function analyzing protein schemas (that bioinformatically define enzymatic consensus sequences) and protein grammar rules that would structurally validate a protein's primary sequence based on \textit{de novo} folding algorithms minimizing a free energy function and/or primary string attribute properties like hydrophobicity, isoelectric charge, amino acid sub-words, \textit{et cetera}. Provided that a basic truthful understanding of protein grammar could be encoded, all evolved proteins in the population could be driven to evolve new functional clades, and eventually protein families. The fitness function could also be coupled with a ``protein spam filter'' that could rule out structurally unstable permutations and combinations, effectively reducing the search space of garbage that needs to be explored (see Figure~\ref{fig:filter}).

To reduce the complexity of designing the initial EASME, the algorithm could drive evolution of very important select proteins, such as key photosynthetic enzymes (for applied use in agriculture); but as EASME matures, any proteins could be evolved and optimized, and all functional islands in the sea of invalidity might be colonized over time.

The final problem EASME will face is understanding how to determine the function of newly evolved proteins derived from \textit{in silico} processes. Our suggestion is that libraries of EASME-derived peptides would first be chemically synthesized, then screened for useful activities. For example, a library of EASME proteins might first be screened against insects, to find EASME-derived proteins capable of killing insects. The positive hits could then be developed into novel insecticides for agriculture, which would also reinforce and improve the grammar structures of the fitness functions. Thus, we outline a feasible path toward the development of the EASME algorithm.

\section{EASME to Date}

While evolutionary methods have been applied to the fields of biology before, EASME represents a specific effort to focus on biologically-accurate representations, with one compelling application being expanding the set of extant proteins. Most EAs applied to biology were abstract top-down population dynamic models. To date, EASME is the first to attempt modeling of actual DNA sequences. For example, the aforementioned cytoplasmic incompatibility research represents an initial development of EASME where both means and ends were represented in biological terms, rather than abstract terms. The individuals being evolved were actual codon sequences, evaluated based on their homology to the real-world \textit{Wolbachia} genome, in order to model their actual evolution. We aim to push the boundaries of genetic programming even further, and use this same methodology to solve problems that have thus far been infeasible to solve \textit{in silico}. While molecular cellular simulations of complete organisms are still computationally intractable, simulating their evolution is feasible.

While other projects have explored this domain in the past (such as ``The optimality of the standard genetic code assessed by an eight-objective evolutionary algorithm,'' by Wnetrzak \textit{et al.} in 2018~\cite{Wnetrzak2018}), they have been relatively few and far-between — and ML has been getting the lion’s share of media attention at the same time. We feel clearly defining EASME as a new field is an important step to clarifying the full potential of CE, and determining where work can be productively focused.

\section{The Path Forward: Other Use Cases for EASME}

\begin{figure}[ht]
    \centering
    \includegraphics[width=.625\textwidth]{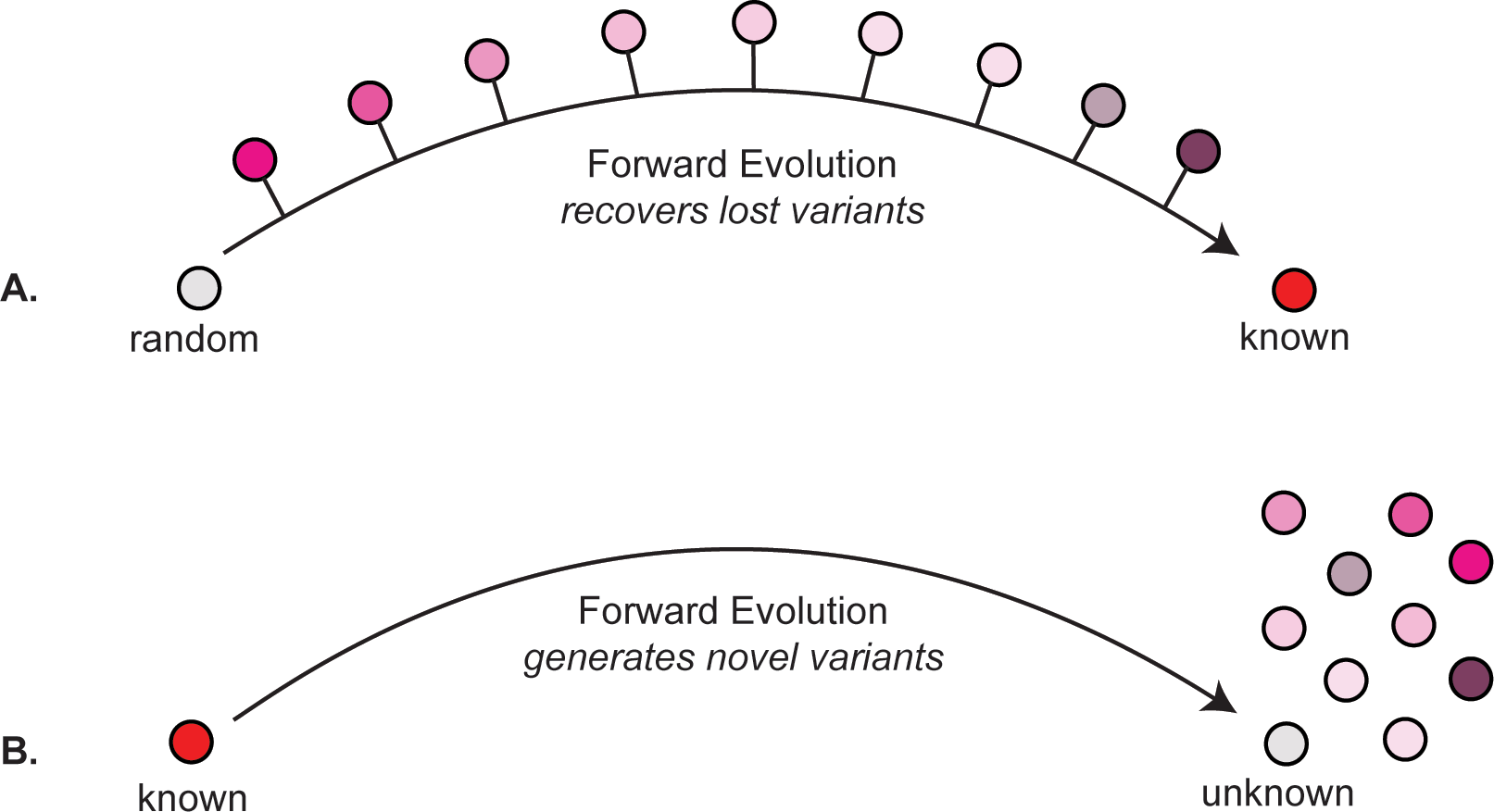}
    \caption{Ways to employ EASME.}
    \label{fig:methods}
\end{figure}

EASME represents a specific effort to focus on expanding the set of extant proteins. The EASME framework, once developed, can run in two distinct ways (see Figure~\ref{fig:methods}). First, EASME can evolve a random sequence toward a known consensus sequence (``\textit{unknown to known}''). In this context, the desired outcome is to reconstruct sequence clusters that went extinct during the process of evolution. Selective fitness is implemented by pushing the evolution towards a known protein sequence family. EASME outputs samples of Pareto optimal sequences from theoretical evolutionary intermediates, effectively recovering extinct sequence variants. How much the EASME generated sequences would differ from real historical intermediates is unknowable without ancient genomes. However, the utility of generated sequences can be tested, measured, and linked to a corresponding successful discovery rate. The second way to run EASME is \textit{known to unknown}, where a known entity is forward evolved into the future by implementing a selection regimen that drives towards a desired characteristic phenotype. This methodology outputs Pareto optimal sequences that may have never evolved yet and is effectively a fast forward button on evolution into the future. While this approach would undoubtedly produce many false positives, wet lab work will allow us to test and validate designed proteins while simultaneously honing a given enzyme's fitness function. Biologically measuring the ratio of valid to invalid protein outputs would allow us to optimize the design process (and even if that ratio is low, it will still be orders of magnitude faster than natural evolution, a process which plays out on evolutionary timescales). To achieve both these ends, EASME will employ EA and GP models supplemented with ML where appropriate. As an aside, EASME can also be run in reverse, where genes and genomes currently extant are reverse-engineered in attempt to reconstruct ancient common ancestors.

Learning the syntax and semantics of our genes is a daunting task, to be sure, but the potential benefits of doing so make the EASME project worth pursuing. With enough time, we can undoubtedly master the biomolecular language. One simple, yet very impactful, possibility is using a proteomic EASME to evolve thermostable variants of existing proteins. The most well-known example of this class of protein is \textit{Taq} polymerase, a thermostable variant of DNA polymerase I, which was discovered in a thermophilic bacteria living in hot springs at Yellowstone National Park~\cite{doi:10.1128/jb.127.3.1550-1557.1976}. This enzyme became the basis for PCR reactions, which amplify DNA, and is used routinely in nearly every biotechnological application. Thermostable proteins evolve by increasing charged amino acid content, which includes insertions of positively-charged lysines (K) and arginines (R), which form thermostable salt bridges with negatively-charged aspartic acid (D) and glutamic acid (E) residues. An EASME algorithm could mutate residues one by one in large populations and (if coupled with \textit{de novo} protein folding and correct protein grammar) might quickly find a Pareto front of the most thermostable enzyme variants for any enzyme known to man. This possibility is currently being explored by the authors. Many enzymatic proteins are used in industrial applications where temperature can fluctuate or needs to be raised, so the potential industrial applications of this project are numerous as well.

One of EASME’s grand challenges is the accurate prediction of protein \textit{functionality}. While projects like AlphaFold have brought protein \textit{structure} prediction closer to reality (albeit with the shortcomings we describe in Section~\ref{sec:ML-shortcomings}), functionality is another matter entirely. EASME, however, may hold the potential to truly solve this problem. A project spearheaded by Wolfgang Banzhaf -- the ``father'' of computational evolution, as described in Section~\ref{sec:EASME-antecedents} -- has recently produced supporting evidence. As detailed in their paper ``A Genetic Programming Approach to Engineering MRI Reporter Genes,'' Banzhaf and his team have been able to use a new protein optimization tool, POET, to design 27 short peptides that produce more contrast in an MRI than was previously state-of-the-art~\cite{doi:10.1021/acssynbio.2c00648}. POET utilizes genetic programming to achieve these results, and while this project could still be considered somewhat narrow in scope (evolving peptides towards one particular function rather than predicting the function of a completely arbitrary peptide), it still demonstrates that evolutionary methods hold potential for solving this problem, and the same methodology could easily be transferred to numerous other functions.

\section{What Needs to Happen Next?}

The authors are currently exploring several promising projects in the field of EASME, as well as the development of a general-purpose “EASME toolkit.” This toolkit will contain the basic elements needed for any EASME project, such as the components of EAs (including genetic programs), the protein spam filter, as well as basic algorithms for determining protein structure, stability, and functionality. Algorithms implemented with this toolkit could be chained together in unique ways (for example, a protein “spam filter” could precede a simulation of protein folding, ensuring computational resources are allocated smartly, allowing for complex projects to be implemented quickly and easily. The authors are also working on an in-depth literature review of related fields, in order to best determine where to focus our algorithmic design efforts, as well as where we can capitalize on existing tools built by others. We hope to publish this survey by the end of this year, with the first version of our toolkit being shared shortly thereafter.

The intersection of computer science and biotechnology is deeply fascinating, and the field has barely been scratched on both sides. While we continue our initial exploratory work, we would highly encourage anyone interested in their own explorations to contact our team. We aim to see EASME reach its full potential, and are certain that great strides will come of this new interdisciplinary methodology.

We plan to document the progress of the EASME field at \url{https://aub.ie/easme}. We welcome contributions from any researchers with an interest in this field.

\section*{Acknowledgements}

The authors gratefully acknowledge Wolfgang Banzhaf's insightful feedback on this paper.

\newpage

\bibliographystyle{apalike}
\bibliography{ref}

\begin{thebibliography}{}

\bibitem[Banzhaf et~al., 2006]{Banzhaf2006}
Banzhaf, W., Beslon, G., Christensen, S., Foster, J.~A., Képès, F., Lefort, V., Miller, J.~F., Radman, M., and Ramsden, J.~J. (2006).
\newblock {From artificial evolution to computational evolution: a research agenda}.
\newblock {\em Nature Reviews - Genetics}, 7:729 -- 735.

\bibitem[Beckmann et~al., 2023]{10.3389/fmicb.2023.1116766}
Beckmann, J., Gillespie, J., and Tauritz, D. (2023).
\newblock {Modeling emergence of Wolbachia toxin-antidote protein functions with an evolutionary algorithm}.
\newblock {\em Frontiers in Microbiology}, 14.

\bibitem[Bricco et~al., 2023]{doi:10.1021/acssynbio.2c00648}
Bricco, A.~R., Miralavy, I., Bo, S., Perlman, O., Korenchan, D.~E., Farrar, C.~T., McMahon, M.~T., Banzhaf, W., and Gilad, A.~A. (2023).
\newblock {A Genetic Programming Approach to Engineering MRI Reporter Genes}.
\newblock {\em ACS Synthetic Biology}, 12(4):1154--1163.
\newblock PMID: 36947694.

\bibitem[Chien et~al., 1976]{doi:10.1128/jb.127.3.1550-1557.1976}
Chien, A., Edgar, D.~B., and Trela, J.~M. (1976).
\newblock Deoxyribonucleic acid polymerase from the extreme thermophile thermus aquaticus.
\newblock {\em Journal of Bacteriology}, 127(3):1550--1557.

\bibitem[D’Angelo et~al., 2023]{DANGELO2023138}
D’Angelo, G., Della-Morte, D., Pastore, D., Donadel, G., {De Stefano}, A., and Palmieri, F. (2023).
\newblock {Identifying patterns in multiple biomarkers to diagnose diabetic foot using an explainable genetic programming-based approach}.
\newblock {\em Future Generation Computer Systems}, 140:138--150.

\bibitem[Robson, 2022]{ROBSON2022105292}
Robson, B. (2022).
\newblock De novo protein folding on computers. benefits and challenges.
\newblock {\em Computers in Biology and Medicine}, 143:105292.

\bibitem[Stevens et~al., 2023]{10.3389/fchem.2023.1106495}
Stevens, J.~A., Grünewald, F., van Tilburg, P. A.~M., König, M., Gilbert, B.~R., Brier, T.~A., Thornburg, Z.~R., Luthey-Schulten, Z., and Marrink, S.~J. (2023).
\newblock Molecular dynamics simulation of an entire cell.
\newblock {\em Frontiers in Chemistry}, 11.

\bibitem[Wnetrzak et~al., 2018]{Wnetrzak2018}
Wnetrzak, M., B{\l}a{\.{z}}ej, P., Mackiewicz, D., and Mackiewicz, P. (2018).
\newblock The optimality of the standard genetic code assessed by an eight-objective evolutionary algorithm.
\newblock {\em BMC Evolutionary Biology}, 18(1):192.

\end{thebibliography}

\end{document}